\documentclass[conference]{IEEEtran}
\usepackage{cite}
\usepackage{amsmath,amssymb,amsfonts}
\usepackage{algorithmic}
\usepackage{graphicx}
\usepackage{epsfig}
\usepackage{textcomp}
\usepackage{xcolor}
\usepackage{float}

\usepackage{times}

\usepackage{caption,subcaption}
\usepackage{float}

\usepackage{hyperref}

\def\BibTeX{{\rm B\kern-.05em{\sc i\kern-.025em b}\kern-.08em
    T\kern-.1667em\lower.7ex\hbox{E}\kern-.125emX}}
\begin{document}

\title{PACE: Posthoc Architecture-Agnostic Concept Extractor for Explaining CNNs \\
}

\author{\IEEEauthorblockN{Vidhya Kamakshi \IEEEauthorrefmark{1}, Uday Gupta \IEEEauthorrefmark{2}, Narayanan C Krishnan \IEEEauthorrefmark{3}}
\IEEEauthorblockA{
\textit{Department of Computer Science and Engineering,} 
\textit{Indian Institute of Technology Ropar,}
Rupnagar - 140001, Punjab, India. \\
Email: \IEEEauthorrefmark{1}2017csz0005@iitrpr.ac.in,
\IEEEauthorrefmark{2}2019csb1127@iitrpr.ac.in,
\IEEEauthorrefmark{3}ckn@iitrpr.ac.in
}
}

\maketitle

\begin{abstract}

Deep CNNs, though have achieved the state of the art performance in image classification tasks, remain a black-box to a human using them. There is a growing interest in explaining the working of these deep models to improve their trustworthiness. In this paper, we introduce a Posthoc Architecture-agnostic Concept Extractor (PACE) that automatically extracts smaller sub-regions of the image called concepts relevant to the black-box prediction. PACE tightly integrates the faithfulness of the explanatory framework to the black-box model. 
To the best of our knowledge, this is the first work that extracts class-specific discriminative concepts in a posthoc manner automatically. The PACE framework is used to generate explanations for two different CNN architectures trained for classifying the AWA2 and Imagenet-Birds datasets. Extensive human subject experiments are conducted to validate the human interpretability and consistency of the explanations extracted by PACE. The results from these experiments suggest that over 72\% of the concepts extracted by PACE are human interpretable.
%To the best of our knowledge, we are the first to extract class-specific discriminative concepts in a posthoc manner automatically. We use the PACE framework to generate explanations for two different CNN architectures trained for classifying the AWA2 and Imagenet-Birds datasets. Further, we conduct extensive human subject experiments to validate the human interpretability and consistency of the explanations extracted by PACE. The results from these experiments suggest that over 72\% of the concepts extracted by PACE are human interpretable.

%There is a hesitation to adopt deep convolutional networks for critical applications due to their black-box nature. Hence there is a growing interest in explaining the working of these deep models that achieve a state of the art performance. We propose PACE, a Posthoc Architecture agnostic Concept Extractor, which extracts smaller sub-regions of the image called concepts relevant to the black-box prediction. Our extensive human subject experiments suggest that the concepts extracted are human interpretable. To the best of our knowledge, we are the first to extract class-specific discriminate concepts in a posthoc manner. We curate a baseline that mimics our explainer's aspects and quantitatively show that our proposed explainer has an upper edge when experimented with different datasets and black-box architectures.

\end{abstract}

\begin{IEEEkeywords}
XAI, posthoc explanations, concept-based explanations, image classifier explanations.
\end{IEEEkeywords}

\section{Introduction}
% Need for explainability
 %Deep Convolutional Neural Network Architectures like VGG \cite{vgg}, and ResNet \cite{resnet} have achieved the state of the art performance on image classification tasks. However, there is a hesitation to adopt these models for safety-critical applications \cite{lipton_2017_doctor_hesitation} due to their black-box nature, i.e., their internal working is not interpretable to the humans using them.  Also, the \textit{Right to Explanation} act by the European Union \cite{EU_right_to_explanation} has made it integral to incorporate Explanations along with decisions of the model, which has lead to a surge in research on \textit{Explainable AI}.
Deep Convolutional Neural Network Architectures like VGG \cite{vgg}, and ResNet \cite{resnet} have achieved a state of the art performance on image classification tasks. However, there is a hesitation to adopt these models for safety-critical applications \cite{lipton_2017_doctor_hesitation} due to their internal working not being interpretable to the humans using them.  Also, the \textit{Right to Explanation} act by the European Union \cite{EU_right_to_explanation} has made it integral to incorporate Explanations along with decisions of the model, which has lead to a surge in research on \textit{Explainable AI}. 

% shortcomings of current approaches
%The working of the black-box image classification network has been explained in different ways. A few approaches modify the input image successively until there is a change in the black-box prediction probability \cite{perturbation_preservation,perturbation_deletion}. But due to the voluminous amount of possible perturbations, these methods cannot be faithfully applied in the real-world scenario. Another family of methods use gradients \cite{gradCAM,grad_cam_plus_plus,full_gradient,vis_cnn} or a specialised attribution score \cite{excitation_bp,score_cam,ablation_cam} propagated across the various layers of the network in order to uncover the salient regions integral to the prediction. However, recent work has shown that faithfulness is not guaranteed when such approaches are applied as they produce similar explanations irrespective of the class label being queried or changes in the underlying black-box model \cite{adebayo2018sanity,sixt_icml_2020}. Further, a single salient region does not provide finer details on the contribution of each of the constituent parts of an image as perceived by humans. 
Early explainability approaches used practically infeasible perturbation strategy \cite{perturbation_deletion,perturbation_preservation} or gradients \cite{gradCAM,grad_cam_plus_plus,full_gradient,vis_cnn} or a specialised attribution score \cite{excitation_bp,score_cam,ablation_cam} propagated to uncover the salient regions integral to the prediction. However, these approaches are not faithful as they produce similar explanations irrespective of the class label being queried or changes in the underlying black-box model \cite{adebayo2018sanity,sixt_icml_2020}. Further, a single salient region does not provide finer details on the contribution of each constituent part of an image as humans perceive the object.

\begin{figure*}
        \begin{subfigure}[b]{0.2\textwidth}
                \includegraphics[width=\linewidth]{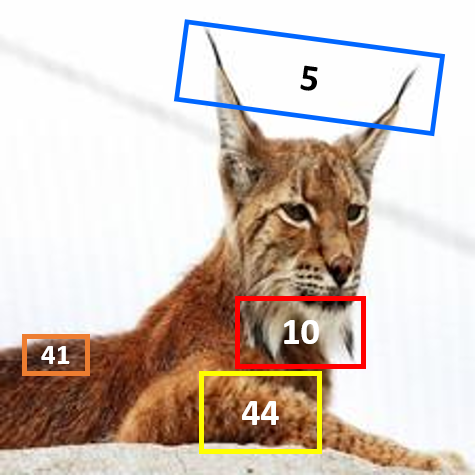}
               \caption{Bobcat}
                \label{fig:bobcat1}
        \end{subfigure}
        \hfill
        \begin{subfigure}[b]{0.2\textwidth}
                \includegraphics[width=\linewidth]{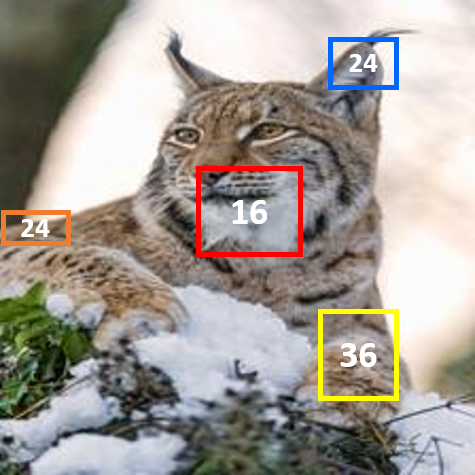}
                \caption{Bobcat}
                \label{fig:bobcat2}
        \end{subfigure}
        \hfill
        \begin{subfigure}[b]{0.2\textwidth}
                \includegraphics[width=\linewidth]{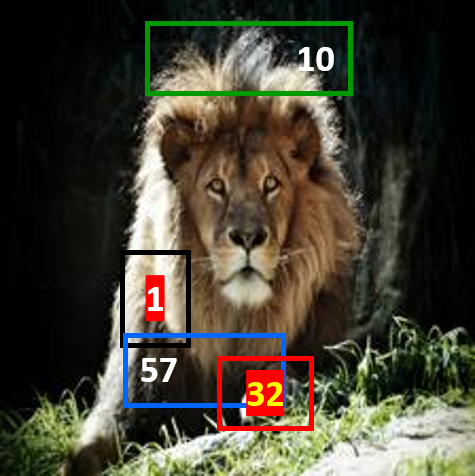}
                \caption{Lion}
                \label{fig:lion1}
        \end{subfigure}
        \hfill
        \begin{subfigure}[b]{0.2\textwidth}
                \includegraphics[width=\linewidth]{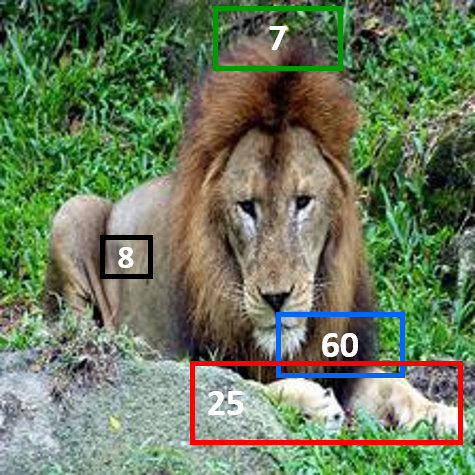}
                \caption{Lion}
                \label{fig:lion2}
        \end{subfigure}
        \hfill
        \begin{subfigure}[b]{0.2\textwidth}
               \includegraphics[width=\linewidth]{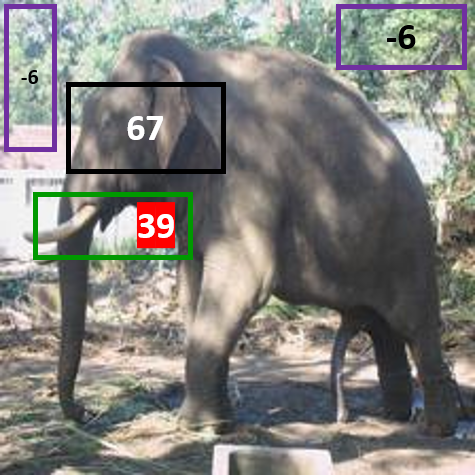}
                \caption{Elephant}
                \label{fig:elephant1}
        \end{subfigure}
        \hfill
        \begin{subfigure}[b]{0.2\textwidth}
                \includegraphics[width=\linewidth]{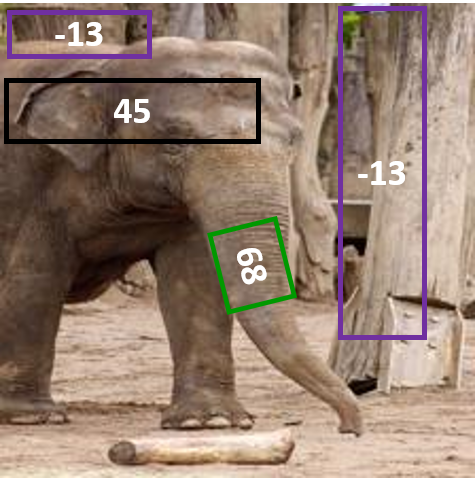}
                \caption{Elephant}
                \label{fig:elephant2}
        \end{subfigure}
        \hfill
        \begin{subfigure}[b]{0.2\textwidth}
             \includegraphics[width=\linewidth]{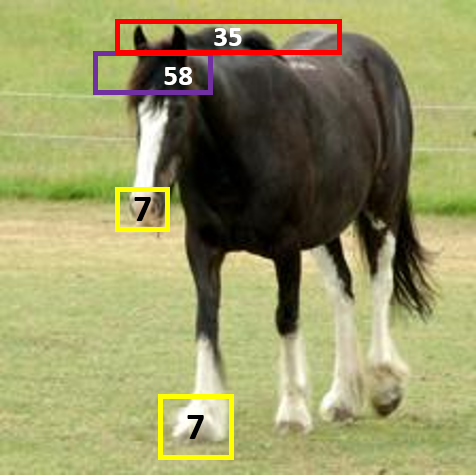}
               \caption{Horse}
              \label{fig:horse1}
        \end{subfigure}
        \hfill
        \begin{subfigure}[b]{0.2\textwidth}
                \includegraphics[width=\linewidth]{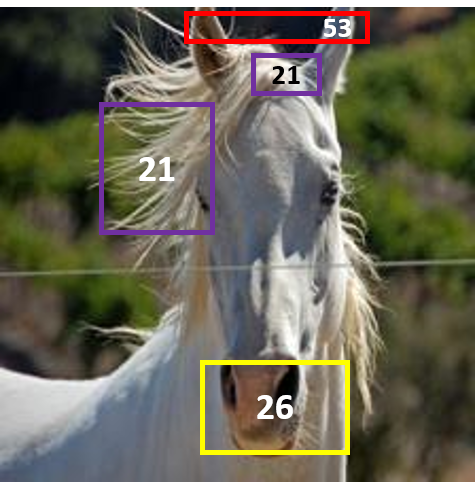}
                \caption{Horse}
                \label{fig:horse2}
        \end{subfigure}
        
        \caption{[Best viewed in color] Class specific concepts extracted by the model from test images of different classes from the AwA2 dataset with their percentage contribution in the box  }\label{fig:various_concepts_test_images}
\end{figure*}

Humans recognize an object through its different salient features \cite{this_looks_like_that,hierarchical_prototypes}. PACE aims to mimic this style of reasoning for explaining the behavior of a black-box image classification model by extracting smaller salient regions in the given image called concepts, which a black-box classifier deems relevant for the prediction. Ideally, a concept can be any human interpretable feature/image-region, say, legs of a lion, stripes of a tiger, body texture of a leopard, background information such as the presence of water, grass, etc. Few concepts extracted from some of the test images are shown in Figure ~\ref{fig:various_concepts_test_images}. As can be seen, the concepts represent salient parts of the different animals such as ears of the bobcat, mane of the lion, trunk of the elephant, mouth of the horse, etc.
%We aim to mimic this style of reasoning for explaining the behavior of a black-box image classification model. We propose to extract smaller salient regions in the given image called concepts, which a black-box classifier deems relevant for the prediction.  Ideally, a concept can be any human interpretable image region, say, legs of a lion, stripes of a tiger, body texture of a leopard, background information such as the presence of water, grass, etc. Few concepts extracted from some of the test images are shown in Figure ~\ref{fig:various_concepts_test_images}. As can be seen, the concepts represent salient parts of the different animals such as ears of the bobcat, mane of the lion, trunk of the elephant, mouth of the horse, etc.

%For example, humans reason about the different parts of an object, such as skin color, texture, facial features, surroundings, etc., to classify the image. 

The proposed framework assumes that every class can be explained by the presence (or absence) of certain characteristics - the concepts. 
The concepts, represented as vectors in a latent space are global, in the sense, they cater to the explanation of a class as a whole. Simultaneously, every input image has different manifestations of the concept vectors - named as embedding vectors. The embedding vectors are extracted through an encoder that works on the feature maps obtained from the black-box. The similarity between the embedding and concept vectors determines the presence of a concept and the visualization. The embeddings are learned such that the output (classification probabilities) of the black-box model for each of the classes is preserved on passing the reconstructed feature map. The relevance of the embedding (and thereby the concept) is obtained by mimicking its removal and observing the drop in the classification probability. This definition of relevance incorporates the faithfulness of the explainer to the black-box by design.
%We represent the concepts as a vector in a latent space. The concept vectors are global, in the sense, they cater to the explanation of a class as a whole. Simultaneously, every input image has different manifestations of the concept vectors - named as embedding vectors. The embedding vectors are extracted through an encoder that works on the feature maps obtained from the black-box. The similarity between the embedding and concept vectors determines the presence of a concept and the visualization. The embeddings are learned such that the output (classification probabilities) of the black-box model for each of the classes is preserved on passing the reconstructed feature map. The relevance of the embedding (and thereby the concept) is obtained by mimicking its removal and observing the drop in the classification probability. This definition of relevance incorporates the faithfulness of the explainer to the black-box by design.

% newly added - vidhya
To explain how a test image has been classified, PACE highlights the salient concepts and provides relevance, denoting the concepts' contribution towards the prediction. The relevance values lie in the range $[-1,1]$. A positive relevance indicates that the concept supports the prediction and a negative denotes that the concept's presence inhibits the prediction. The relevances are normalized, and the percentage contribution of the different concepts towards the prediction of various test images has also been shown in Figure ~\ref{fig:various_concepts_test_images}. For instance, consider the elephant's image shown in Figure ~\ref{fig:elephant1}. The concept face has a contribution of 67\%; the trunk has a contribution of 39\%. These concepts support the prediction of the image as an elephant. At the same time, the concept of trees has a negative contribution (-6\%). This can be understood as trees may be present in the background of different animals. Hence, the presence of trees may not support the prediction of the animal. Due to the presence of trunk and face that strongly supports the animal being predicted as an elephant, the given test image was predicted as an elephant.
Overall, the major contributions of the proposed work are:
\begin{itemize}
    \item To the best of our knowledge, this is the first work that extracts relevant and discriminative class-specific concepts to explain the behavior of any black-box CNN.
    \item The approach tightly integrates the relevant concept extraction into the explanation learning process, instead of leaving it as a post-training step.
    \item Extensive human-subject experiments are conducted to validate the consistency and interpretability of the concepts.
\end{itemize}

\section{Related Work}

    A lot of work has gone in to explain the output of the black-box networks. Broadly, these approaches can be categorized into Antehoc and Posthoc methods. 
    
    Antehoc methods, also referred to as Explainable by Design methods, incorporate explainability into the model during the training phase itself. These approaches may require changes to the architecture and retraining to explain the working of an already deployed black-box model. Class Activation Maps (CAM) \cite{cam} is one of the earliest approaches that performs architecture modification and retraining to incorporate the explainability aspect. Li et al. \cite{aaai_2018_prototype_vector} propose an autoencoder based architecture that incorporates explanations in terms of proximity to characteristic prototypes, which is then used to perform classification. This looks like That paradigm \cite{this_looks_like_that} proposes using a convolutional encoder to learn class-specific prototypes, which are then linearly combined to perform classification. Hase et al.\cite{hierarchical_prototypes} leverage the work of Chen et al. \cite{this_looks_like_that} to perform hierarchical classification by incorporating explainability in their design to learn class-discriminate prototypes at each level of the hierarchy. 
    
    Posthoc methods do not require any architecture modification or black-box retraining. They probe the trained black-box model to understand its working. The initial work followed a perturbation based strategy \cite{perturbation_deletion,perturbation_preservation} where the image is successively edited until a significant change in the prediction probability is observed. The feature whose perturbation causes a significant change is deemed integral for the prediction.
    However, due to the voluminous amount of possible perturbations, it is not guaranteed that a non-brute force, heuristic based perturbation approach can figure out the features that are integral to the black-box prediction in a faithful manner. 
    
    On the other hand, saliency-based Posthoc methods investigate the internals of the network to identify the region relevant for the black-box model to make the prediction. Grad-CAM \cite{gradCAM} is the generalization of CAM \cite{cam} that leverages gradients flowing into the final convolutional layers to localize salient region. Grad CAM++ \cite{grad_cam_plus_plus} helps localize multiple occurrences of the same object but requires the computation of higher-order derivatives. Full Grad \cite{full_gradient} utilizes both bias and input gradients to achieve pixel-level attribution. However, recent literature suggests a possible compromise in the faithfulness of the explanations when gradients are used \cite{adebayo2018sanity,sixt_icml_2020,similar_saliency_ieee}. Other recent approaches like Score CAM \cite{score_cam}, Ablation CAM \cite{ablation_cam}, and Eigen CAM \cite{eigen_cam} are variations of CAM that do not use gradients to localize the salient region. However, these approaches output only a heatmap where a single blob in the image is highlighted but does not reveal smaller regions' relevances in the image.
    
    On the other hand, a few posthoc approaches \cite{lime,shap, anchors,maire,muse} aim to construct an interpretable approximation to explain the working of any black-box model, not restricted to CNNs. Ribeiro et al. \cite{lime} learn piece-wise locally linear approximations to explain the working of any complex function learned by the black-box model. Ribeiro et al. \cite{anchors} proposes constructing a subset of features integral to the predictions in a bottom-up fashion called Anchors as generalizable explanations. Lundberg \& Lee \cite{shap} use Game-theoretic Shapley values to quantify the relevance of each feature towards prediction. MAIRE  \cite{maire} extends Anchors \cite{anchors} to be applied on continuous data without the need for binning or discretization by formulating to find the optimal orthotope that explains the prediction of a given test instance. MUSE\cite{muse} also provides explanations similar to that of MAIRE \cite{maire} but requires the user to input the value ranges of features at which the explanation should be generated. The need to extract super-pixels from the images to explain the classification output is a fundamental limitation of all these approaches.
    
    A new class of approaches aims to explain the working of the black-box through human interpretable concepts, which are vectors in the latent activation space. TCAV \cite{tcav} requires the users to provide examples of concepts, while ACE \cite{ghorbani_tace_nips_2019} uses segmentation to automatically extract the concepts. The limitation of these approaches is finding the relevance of a concept using directional derivatives, which is a weaker (linear) approximation, given the network's non-linearity. Wu et al. \cite{global_concept_attribution} propose to extract the contribution of each concept towards the prediction in a similar manner as TCAV \cite{tcav} and ACE \cite{ghorbani_tace_nips_2019} by leveraging directional derivatives. However, the visualization of the manifestation of the concept requires Activation Maximization (AM) \cite{dgn_am} techniques, which makes it difficult to explain to the users who have good domain knowledge but little to no deep learning expertise. Concept SHAP \cite{concept_shap} extracts concepts in an unsupervised manner without the image segment assumption of concepts whose relevance is quantified utilizing Shapley values. However, a two-layer non-linear network is involved in Concept SHAP to learn the concept embeddings, which leads to using another black-box to explain the given black-box. ICE \cite{ice} learns integral concepts using Non-negative Matrix Factorization. While these approaches learn generic concepts for the whole dataset, the proposed work aims to learn class-specific concepts to improve the explanations' interpretability.

\begin{figure*}
    \centering
    \includegraphics[width =  \textwidth]{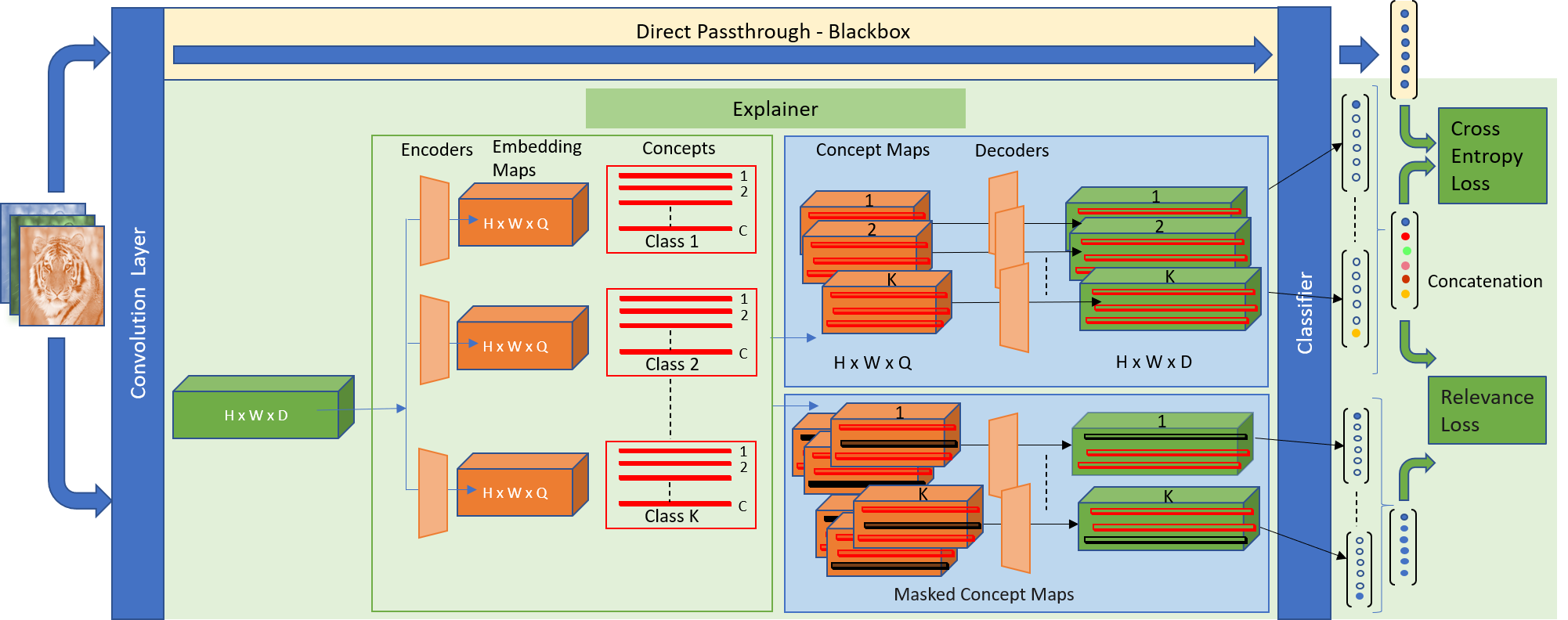}
    \caption{[Best  viewed in color] Various modules in the proposed PACE framework}
    \label{fig:explainer_architecture}
\end{figure*}

\section{Methodology}
%We propose PACE which is capable of explaining the working of any convolutional layer. Ours is a grey box explainability approach in the sense that we just need the feature map from the layer of interest and the probability distribution which is the output of the black-box to generate explanations. We aim to learn class specific concepts that are integral towards the black-box predictions.

%We propose PACE that can be used to explain the working of any convolutional layer by learning class discriminant regions of the image which we call concepts along with their relevance towards prediction. The PACE framework breaks open the network at the convolutional layer of interest and helps peep the working of the black-box through the lens of various concepts detected at that convolutional layer of interest. 
        % --> 2 lines merged to 1 lin
        
% explainer architecture here

The proposed PACE framework dissects the convolutional layer of a black-box model to uncover latent representations of class discriminative image regions. Figure ~\ref{fig:explainer_architecture} presents the schematic diagram of the framework, as well as illustrates the two primary components, namely, an autoencoder (AE) and the global concept representations (concepts). The encoder part of the AE transforms the convolutional feature map of an input image into a representation in the space of concepts, while the decoder part of the AE projects the vectors in the latent concept space back to the space of the convolutional feature map. The search for the presence of the concepts happens in the latent concept space. 
%On the other hand, the decoder part of the AE projects the vectors in the latent concept space back to the space of the convolutional feature map. The search for the presence of the global concepts happens in the latent concept space.

The encoder is designed as a $1-D$ convolutional layer that aims to project the feature map representation onto a low dimensional ($Q$) embedding concept space. The encoder's goal is to coalesce the information pertaining to different concepts spread across different feature maps into a more compact representation. The encoder is linear and thus retains interpretability - the concept representations may be interpreted as weighted combinations of the input features. Other approaches like Concept-SHAP \cite{concept_shap} use non-linear activations, thereby reducing the explanation framework's interpretability. The decoder in the PACE framework is also designed to be a linear transpose convolutional layer transforming the vectors in the latent space to the feature maps.

Each class $k$ is represented by a set of $C$ concepts denoted by $\mathcal{C}_k$ such that the latent representation of the same(different) concept of a class are similar(dissimilar) across different instances of that class. To explain a $K$-way classifier, PACE leverages $K$ independent autoencoders, each dedicated for a class. The feature maps $F \in \mathbb{R}^{H \times W \times D}$ from the convolutional layer of interest are passed through each of the $K$ autoencoders. $H$ and $W$ denote the feature maps' height and width, and $D$ denotes the number of channels. The $k^{th}$ autoencoder (parameterized by $\theta^k$) is trained independently to learn concepts related only to the $k^{th}$ class. The latent space for every autoencoder is different, though the dimensionality is the same.  
%The concept representations lie in the latent space of the AE. To explain a $K$-way classifier, PACE leverages $K$ independent autoencoders, each dedicated for a class. The feature maps $F \in \mathbb{R}^{H \times W \times D}$ from the convolutional layer of interest are passed through each of the $K$ autoencoders. $H$ and $W$ denote the feature maps' height and width, and $D$ denotes the number of channels. The $k^{th}$ autoencoder (parameterized by $\theta^k$) is trained independently to learn concepts related only to the $k^{th}$ class. The latent space for every autoencoder is different, though the dimensionality is the same. 

The encoder's output for an input image $\textbf{x}$ is an embedding map ($E_k \in \mathbb{R}^{H \times W \times Q}$). The embedding map $E_k$ denotes the concepts' manifestation at each of the $H\times W$ locations in the feature map. Once the latent concept vectors are learned (the learning procedure will be explained later), the similarity of the $Q$-dimensional embedding vector at each of $H\times W$ locations with respect to the concept vectors for class $k$ can be determined. This results in $C$ similarity matrices denoted by $S_k$, each of dimension $H\times W$. We use the inverse of the Euclidean distance between the embedding vector and the concept vector as the similarity measure. A concept $\mathcal{C}_k^j$ is present in the feature map at the spatial location $(l,m)$ if $S_k^j[l,m]$ exceeds a threshold $\tau$ determined relative to the maximum value. The similarity matrices can be treated as masks to visualize the concepts after suitable resizing.

%The encoder's output for an input image $\textbf{x}$ is an embedding map ($E_k \in \mathbb{R}^{H \times W \times Q}$). The embedding map $E_k$ denotes the concepts' manifestation at each of the $H\times W$ locations in the feature map. Let us assume that the global concept vectors in this $Q$-dimensional latent space have been learned (later, we explain how to learn the global concepts). We can determine the similarity of the $Q$-dimensional embedding vector at each of $H\times W$ locations with respect to the global concept vectors for class $k$. This results in $C$ similarity matrices denoted by $S_k$, each of dimension $H\times W$. We use the inverse of the Euclidean distance between the embedding vector and the global concept vector as the similarity measure. A global concept $\mathcal{C}_k^j$ is present in the feature map at the spatial location $(l,m)$ if $S_k^j[l,m]$ exceeds a threshold $\tau$ determined relative to the maximum value. The similarity matrices can be treated as masks to visualize the concepts after suitable resizing.

The decoder (parameterized by $\phi^k$) of the AE for class $k$ works on the embedding map $E_k$ to reconstruct the original feature map $F$.The concept vector should lie in the embedding manifold. Then replacing the embedding vector in $E_k$ with the most similar concept vector at locations with the strong presence of the concept should not alter the decoder's output. This idea is used to enforce alignment between the concept vectors and the embedding manifolds, thus assisting in learning the concept vectors. Specifically, the embedding vector at a spatial location $(l,m)$ is replaced with the most similar global concept vector $\mathcal{C}^{j}_{k} \in \mathbb{R}^Q$ if the concept $j$ is strongly present at that spatial location. This gives the Concept Map $\tilde{E_k} \in \mathbb{R}^{H \times W \times Q}$, which is then passed through the decoder to obtain the reconstructed feature map $\hat{F_k}$ corresponding to the class $k$ module. 
%The global concept vector should lie in the embedding manifold. Then replacing the embedding vector in $E_k$ with the most similar global concept vector at locations with the strong presence of the concept should not alter the decoder's output. This idea is used to enforce alignment between the global concept vector and the embedding manifolds, thus assisting in learning the global concept vectors. Specifically, we replace the embedding vector at a spatial location $(l,m)$ with the most similar global concept vector $\mathcal{C}^{j}_{k} \in \mathbb{R}^Q$ if the concept $j$ is strongly present at that spatial location. This gives us the Concept Map $\tilde{E_k} \in \mathbb{R}^{H \times W \times Q}$, which is then passed through the decoder to obtain the reconstructed feature map $\hat{F_k}$ corresponding to the class $k$ module. 

The reconstructed feature map, $\hat{F_k}$, is then passed through the rest of the black-box to get the prediction probabilities. Let $p_k$ represent the prediction probability obtained for class $k$ using $\hat{F_k}$, and $P$ the concatenation of the corresponding class probabilities obtained from all the $K$ reconstructed feature maps. Each autoencoder ($\theta^k$, $\phi^k$) learns to detect concepts that are integral only for class $k$, therefore is only reliable in explaining the output of the black-box model for class $k$. According to the PACE explainer, the class label with the highest probability in $P$ is the predicted label. 

%As discussed before, if the embedding and global concept vectors are close, then $P$ obtained via the reconstructed feature maps $\hat{F_k}$ should be similar to the classification probabilities obtained from the original feature map $F$. This is enforced by using a Cross-Entropy loss between $P$ and the black-box prediction $b(\textbf{x})$ defined as
As discussed before, if the embedding and concept vectors are close, then $P$ obtained via the reconstructed feature maps $\hat{F_k}$ should be similar to the classification probabilities obtained from the original feature map $F$. This is enforced by using a Cross-Entropy loss between $P$ and the black-box prediction $b(\textbf{x})$ defined as
\begin{equation}
    \mathcal{L}_{C} = CrsEnt(P,b(\textbf{x}))
\label{eq:ce_loss}    
\end{equation}
%As a result, even if the manifold of the randomly initialized global concept vectors is not aligned with the embedding manifold, minimizing the above loss will eventually bring them closer.
As a result, even if the manifold of the randomly initialized concept vectors is not aligned with the embedding manifold, minimizing the above loss will eventually bring them closer. 

Further, to ensure that the concept vectors are different from each other, the pairwise Euclidean distance between these vectors of a single class is maximized as given below 
%we maximise the pairwise Euclidean distance between these vectors of a single class as given below
\begin{equation}
    \mathcal{L}_{D} = \sum_{k = 1}^{K} \sum_{j = 1}^{C} \sum_{j^{'} = 1}^{C} ||\mathcal{C}^{j}_{k} - \mathcal{C}^{j^{'}}_{k}||_2^2
\label{eq:distance_loss}
\end{equation}

%The process of extracting distinct global concept vectors is reinforced by applying the triplet loss on the corresponding most similar embedding vectors. Specifically, for the instance $\textbf{x}_i$ in a batch of $B$ images, we obtain the embedding vector $E_k^j(i)$ that is most similar to the global concept $\mathcal{C}_{k}^{j}$. The embedding vectors most similar to the global concept $\mathcal{C}_{k}^{j}$ obtained from the other images in the batch belonging to class $k$ form the set of anchor positives $\mathcal{P}_k^j(i)$. Similarly, the embedding vectors most similar to the other global concept vectors $\mathcal{C}_{k}^{j'\neq j}$ from the images in the batch belonging to class $k$ form the set of anchor negatives $\mathcal{N}_k^j(i)$. We use all anchor-positive pairs and select semi-hard negatives for anchor-negative pairs as suggested by \cite{schroff2015facenet}. The margin $\alpha$ is set to 1 so as to encourage orthogonal embeddings. The triplet loss is thus defined as
The process of extracting distinct concept vectors is reinforced by applying the triplet loss on the corresponding most similar embedding vectors. Specifically, for the instance $\textbf{x}_i$ in a batch of $B$ images, we obtain the embedding vector $E_k^j(i)$ that is most similar to the concept $\mathcal{C}_{k}^{j}$. The embedding vectors most similar to the concept $\mathcal{C}_{k}^{j}$ obtained from the other images in the batch belonging to class $k$ form the set of anchor positives $\mathcal{P}_k^j(i)$. Similarly, the embedding vectors most similar to the other concept vectors $\mathcal{C}_{k}^{j'\neq j}$ from the images in the batch belonging to class $k$ form the set of anchor negatives $\mathcal{N}_k^j(i)$. We use all anchor-positive pairs and select semi-hard negatives for anchor-negative pairs as suggested by \cite{schroff2015facenet}. The margin $\alpha$ is set to 1 so as to encourage orthogonal embeddings. The triplet loss is thus defined as
\begin{equation}
    \mathcal{L}_{T} = \sum_{e_{p} \in \mathcal{P}_k^j(i)} \sum_{e_{n} \in \mathcal{N}_k^j(i)}  ||E_k^j(i) - e_p||_2^2 - ||E_k^j(i) - e_n||_2^2 +\alpha 
\label{eq:triplet_loss}
\end{equation}

The triplet loss requires a sufficient number of anchor positives to learn a good separation \cite{schroff2015facenet}. To ensure this, the training strategy uses a mix of pure and mixed batch instances. A batch is pure if all batch instances are predicted to be of the same class by the black-box; otherwise, it is a mixed batch. It is to be noted that pure batches' formation is based on predicted label (output from the black-box CNN) and not the ground truth. This is because we want the explainer to learn the functioning of the black-box. A single iteration succeeds every $\rho$ number of training iterations involving pure batches over a mixed batch. This helps to learn the interplay of concepts across different classes.

%We mimic the removal of the concept and observe the drop in prediction probability to estimate the relevance of the concept. Specifically, the relevance  $r_k^j \in [-1,1]$  for concept $\mathcal{C}^{j}_{k} \in \mathcal{C}_k$ is obtained in the following manner. We force $\tilde{E_k}[l,m] = \mathbf{0}$ at all spatial locations $(l,m)$ where $\mathcal{C}^{j}_{k}$ is present, resulting in a masked concept map $M_k^j  \in \mathbb{R}^{H \times W \times Q}$. $M_k^j$ is passed through the decoder $\phi^k$ to get the reconstructed feature map (where the concept is removed) and obtain the final classification probability for class $k$, $p_k^j$. Relevance is then computed as the difference in the probabilities. i.e. $r_k^j = p_k - p_k^j$. A positive relevance value denotes that the concept supports in the prediction of class $k$, while a negative relevance value denotes that the concept inhibits the prediction of class $k$. A novel feature of our approach is the ability to learn concepts relevant for the prediction. We achieve this by applying the Squared Error loss between the relevance and the explainer probability defined as
A concept's relevance is estimated by mimicking its removal and observing the drop in prediction probability. Specifically, the relevance  $r_k^j \in [-1,1]$  for concept $\mathcal{C}^{j}_{k} \in \mathcal{C}_k$ is obtained in the following manner. At all spatial locations $(l,m)$ where $\mathcal{C}^{j}_{k}$ is present, $\tilde{E_k}[l,m]$ is forced to be $= \mathbf{0}$, resulting in a masked concept map $M_k^j  \in \mathbb{R}^{H \times W \times Q}$. $M_k^j$ is passed through the decoder $\phi^k$ to get the reconstructed feature map (where the concept is removed) and the final classification probability for class $k$, $p_k^j$ is obtained. Relevance is then computed as the difference in the probabilities. i.e. $r_k^j = p_k - p_k^j$. A positive relevance value denotes that the concept supports in the prediction of class $k$, while a negative relevance value denotes that the concept inhibits the prediction of class $k$. Concepts relevant for the prediction are learned by applying the Squared Error loss between the relevance and the explainer probability defined as
\begin{equation}
    \mathcal{L}_{R} =  \sum_{k = 1} ^{K} \sum_{j = 1} ^{C} ||r^j_k - p_k||^2_2
\label{eq:relevance_loss}
\end{equation}

Thus, the overall loss for training the PACE framework is the weighted combination of these four losses defined as 
\begin{equation}
    \mathcal{L} = \beta\mathcal{L}_C + \gamma\mathcal{L}_R - \delta\mathcal{L}_D + \omega \sum_{i=1}^{B} \sum_{k=1}^{K} \sum_{j=1}^{C} \mathcal{L}_T(i,j,k)
\label{eq:explainer_objective}
\end{equation}
This results in an end-to-end training of the PACE framework for learning $\{(\theta^k, \phi^k)\}_{k=1}^K$ and $\{\mathcal{C}_k\}_{k=1}^K$

\section{Experiments}
%We demonstrate the PACE framework for explaining image classifiers trained on two different datasets - Animals With Attributes 2 (AWA2) \cite{AWA2,awa2_paper_xian_2018_zero}, Imagenet-Birds \cite{imagenet}. We take a subset of 20 classes out of the total 50 classes in AWA2 such that each class has at least 500 images. We take a subset of 10 bird classes from the 1000-way Imagenet dataset \cite{imagenet} as our Imagenet-Birds dataset.

%We demonstrate the PACE framework for explaining image classifiers trained on two different datasets - Animals With Attributes 2 (AWA2) \cite{AWA2,awa2_paper_xian_2018_zero}, Imagenet-Birds \cite{imagenet}.
The PACE framework is used to explain image classifiers trained on two different datasets - Animals With Attributes 2 (AWA2) \cite{awa2_paper_xian_2018_zero}, Imagenet-Birds \cite{imagenet}.
A subset of 20 classes was taken from the 50-way AWA2 dataset.  A subset of 10 classes was taken from the 1000-way Imagenet dataset \cite{imagenet} to build the Imagenet-Birds dataset. 
%We take a subset of 20 classes from the 50-way AWA2 dataset, and 10 bird classes from the 1000-way Imagenet dataset \cite{imagenet} as our Imagenet-Birds dataset.

The PACE framework was used
%We use the PACE framework
to explain the behavior of two different CNN architectures, namely, VGG16 and VGG19. These models were pretrained on the ImageNet dataset \cite{imagenet} and fine-tuned on the corresponding datasets of interest with a train, validation, test split of 80\%, 10\%, 10\% respectively. Adam optimizer \cite{adam} is used to perform optimization in all our experiments. In all the classifier fine-tuning setup, the batch size was 64; the learning rate is $10^{-3}$, and the regularization weight decay parameter is $5 \times 10^{-5}$. The test accuracy on the AWA2(VGG16), Imagenet-Birds(VGG16), and Imagenet-Birds(VGG19) are 92.9\%, 96.6\%, and 97.1\% respectively.

% % Include this in the text corresponding to black box training
% \begin{table}
%     \centering
%     \begin{tabular}{|c|c|c|c|c|c|}
%     \hline
%     Black-box & Dataset & \# epochs  & Test Accuracy\\
%     \hline
%      VGG16 & AWA2 & 30   & 92.9\% \\
%      \hline
%      VGG16 & Imagenet (Birds) & 50   & 96.6\% \\ 
%      \hline
%      VGG19 & Imagenet (Birds) & 100 & 97.1\% \\ 
%      \hline
%     \end{tabular}
%     \caption{Classifier Performance}
%     \label{tab:classifier_performance}
% \end{table}

The PACE explainers for the three models are trained for 100 epochs with a batch size 32, learning rate of $10^{-4}$ and the regularization weight decay parameter is 0.1. The values of the other hyper-parameters for PACE are $C=10$, $Q=32$, $\tau = 95 \%$, $\rho = 5$, $\beta = 100$, $\gamma = 1000$, $\delta = \omega = 1$ obtained via cross validation.

\begin{table}[H]
    \centering
    \begin{tabular}{|c|c|c|c|c|c|}
    \hline
    Black-box & Dataset & PACE (Ours) & Baseline \\
    \hline
     VGG16 & AWA2   & \textbf{88.2\%}  &  51.4\%\\
     \hline
     VGG16 & Imagenet (Birds)  & \textbf{94.7\%}  & 67.3\%\\ 
     \hline
     VGG19 & Imagenet(Birds)  & \textbf{94.1\%}  & 70\%\\           
     \hline
    \end{tabular}
    \caption{Explainer Agreement Accuracies}
    \label{tab:explainer_agreement_accuracy}
\end{table}

\subsection{Comparison with PCA+Clustering Baseline}
A strong baseline to compare our approach would be to cluster the representations obtained after applying PCA on the feature maps. The PCA replicates the linearity of the autoencoder learned by our model, and the clustering (K-means) represents the application of the triplet loss used to learn distinct concepts by the PACE framework. However, this baseline cannot automatically learn class-wise concepts, unlike PACE. This is overcome by explicitly learning the cluster centroids for each class independently using pure batches. Specifically, given a pure batch containing the images for class $k$, a low dimensional embedding map $E_{k}^{B}$ is obtained via PCA from the feature map $F$. These embeddings are clustered to get $C$ clusters representing concept vectors for that class $k$ denoted by $\mathcal{C}_{k}^{B}$. The low dimensional embedding map $E_{k}^{B}$ (with the embedding vector replaced by the most similar cluster centroid) can be transformed to obtain the approximation to the feature map $F$, which in turn can be used to obtain the classification probabilities.
%We overcome this shortcoming by explicitly learning the cluster centroids for each class independently using pure batches. Specifically, given a pure batch containing the images for class $k$, we perform PCA to get a low dimensional embedding map $E_{k}^{B}$ from the feature map $F$. These embeddings are clustered to get $C$ clusters. 
%The $C$ centroids are taken as representative global concept vectors for that class $k$ denoted by $\mathcal{C}_{k}^{B}$. The low dimensional embedding map $E_{k}^{B}$ (with the embedding vector replaced by the most similar cluster centroid) can be transformed to obtain the approximation to the feature map $F$, which in turn can be used to obtain the classification probabilities.

%We measure the agreement between the predicted labels of the baseline and the black-box model. Similarly, we also measure the agreement in the PACE framework's predicted labels and the black-box model. These scores for the three CNN models are presented in Table ~\ref{tab:explainer_agreement_accuracy}. It can be seen that PACE significantly outperforms the baseline in all the cases. We do not include the baseline for the human subject experiments due to the low agreement accuracy. 
The \% of test instances where the label as predicted by the explainer ($arg\max_k p_k(x)$) and the black-box ($arg\max_k b_k(x)$) agree is termed the agreement accuracy. These scores for the three CNN models are presented in Table ~\ref{tab:explainer_agreement_accuracy}. It can be seen that PACE significantly outperforms the baseline in all the cases. The baseline is not included for the human subject experiments due to the low agreement accuracy.

%human subject survey
\subsection{Human Subject Experiments}
%We conduct human subject experiments to assess the interpretability and consistency of the class-specific concepts extracted by the PACE framework. A concept-tagging experiment involving 100 subjects was conducted using the concepts extracted by the PACE framework on the VGG16 model trained for the AWA2 dataset. Every participant was asked 20 unique questions (10 classes $\times$ 2 concepts per class). In each question, pertaining to a single concept, the participant was presented with five different images from the same class having the visualization of the concept. The participants were asked if they could observe any common pattern across the five visualizations and, if so,  were also asked to tag the concept. Two randomly selected questions were duplicated to validate the consistency of the responses of the individual participants. 
Human subject experiments were conducted to assess the interpretability and consistency of the class-specific concepts extracted by the PACE framework.  A concept-tagging experiment involving 100 subjects was conducted using the concepts extracted by the PACE framework on the VGG16 model trained for the AWA2 dataset. Every participant was asked 20 unique questions (10 classes $\times$ 2 concepts per class). In each question, pertaining to a single concept, the participant was presented with five different images from the same class having the visualization of the concept. The participants were asked if they could observe any common pattern across the five visualizations and, if so,  were also asked to tag the concept. Two randomly selected questions were duplicated to validate the consistency of the responses of the individual participants. 

\begin{figure}
    \centering
    \includegraphics[width = 0.75\linewidth]{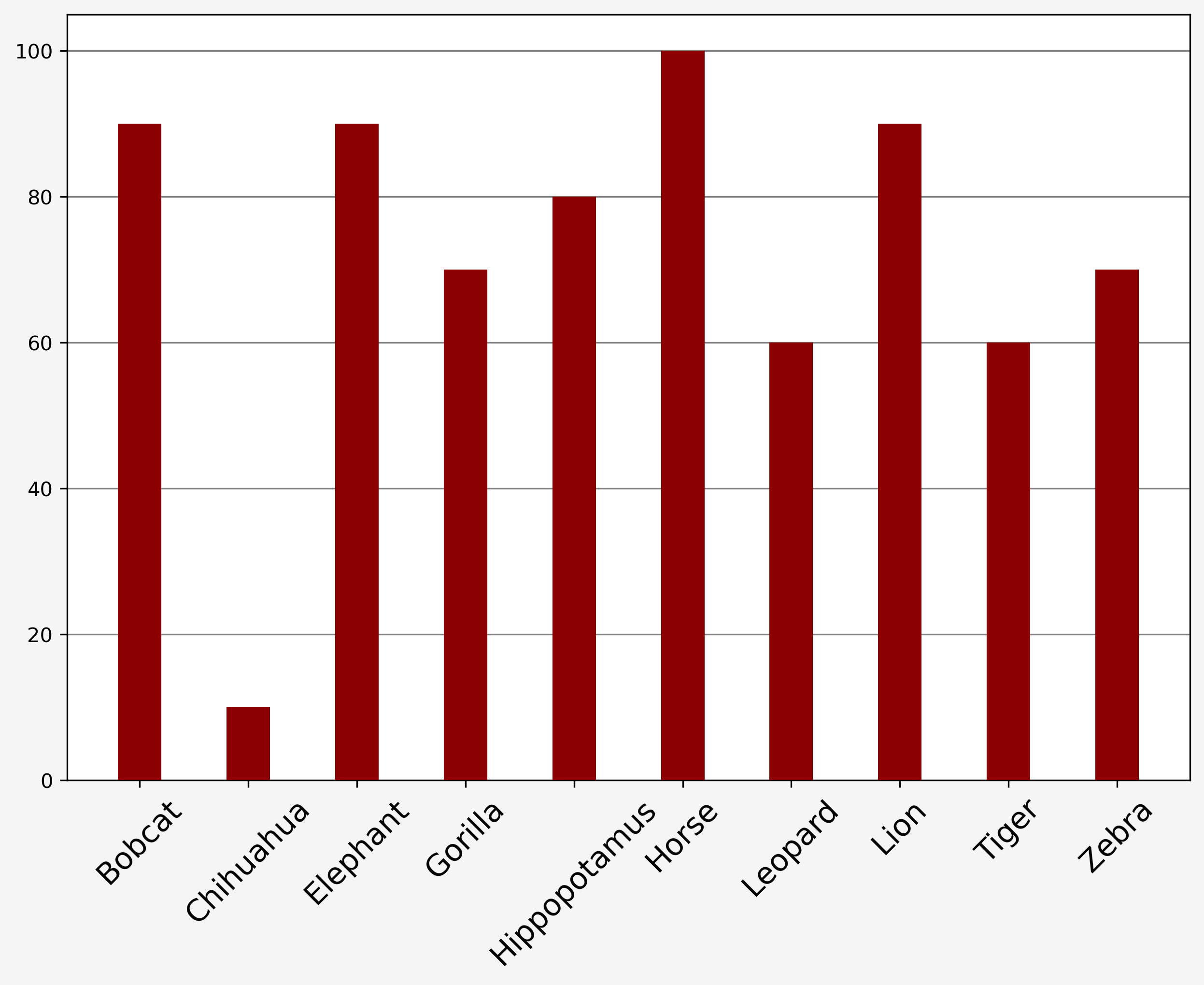}
    \caption{Human interpretable concepts per class (percentage) 
}
    \label{fig:survey_bargraph}
\end{figure}
% reason for low consistency of chihuahua - added - vidhya

\begin{figure}
    %\centering
    \begin{subfigure} {0.5\textwidth}
            \includegraphics[width = 0.18 \linewidth]{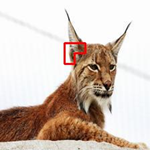}
            %\hfill
            \includegraphics[width = 0.18 \linewidth]{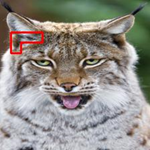}
            %\hfill
            \includegraphics[width = 0.18 \linewidth]{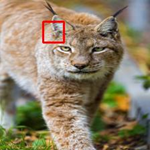}
            %\hfill
            \includegraphics[width = 0.18 \linewidth]{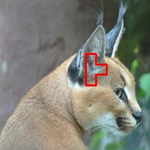}
            %\hfill
            \includegraphics[width = 0.18 \linewidth]{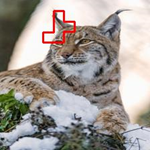}
            \caption{Bobcat - Ear, Right ear, Ear hair, Ear structure}
            \label{fig:bobcat_ear}
    \end{subfigure}
    \begin{subfigure} {0.5\textwidth}
        \includegraphics[width = 0.18 \linewidth]{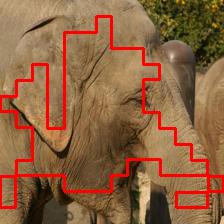}
        %\hfill
        \includegraphics[width = 0.18 \linewidth]{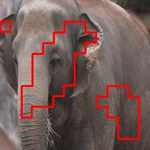}
        %\hfill
        \includegraphics[width = 0.18 \linewidth]{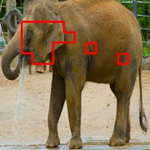}
        %\hfill
        \includegraphics[width = 0.18 \linewidth]{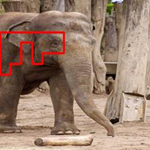}
        %\hfill
        \includegraphics[width = 0.18 \linewidth]{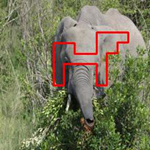}
        \caption{Elephant - Face, Eye, Ear}
        \label{fig:elephant_face}
    \end{subfigure}
    \begin{subfigure} {0.5\textwidth}
        \includegraphics[width = 0.18 \linewidth]{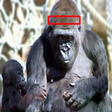}
        %\hfill
        \includegraphics[width = 0.18 \linewidth]{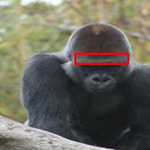}
        %\hfill
        \includegraphics[width = 0.18 \linewidth]{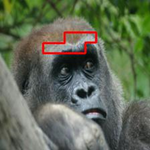}
        %\hfill
        \includegraphics[width = 0.18 \linewidth]{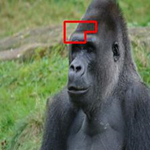}
        %\hfill
        \includegraphics[width = 0.18 \linewidth]{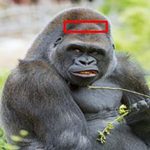}
        %\hfill
        \caption{Gorilla - Forehead, Head, Hair}
        \label{fig:gorilla_forehead}
    \end{subfigure}
    \begin{subfigure}{0.5\textwidth}
        \includegraphics[width= 0.18 \linewidth]{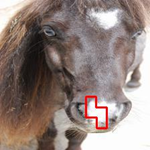}
        %\hfill
        \includegraphics[width= 0.18 \linewidth]{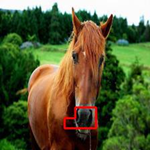}
        %\hfill
        \includegraphics[width= 0.18 \linewidth]{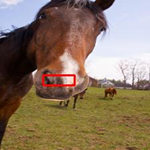}
        %\hfill
        \includegraphics[width= 0.18 \linewidth]{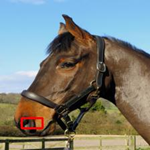}
        %\hfill
        \includegraphics[width= 0.18 \linewidth]{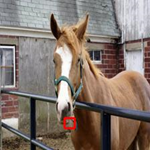}
        \caption{Horse - Muzzle, Mouth, Nose, Nostrils, Snout}
        \label{fig:horse_muzzle}
    \end{subfigure}
    \begin{subfigure} {0.5\textwidth}
        \includegraphics[width = 0.18 \linewidth]{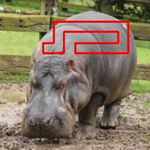}
        %\hfill
        \includegraphics[width = 0.18 \linewidth]{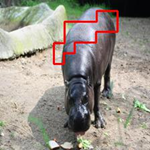}
        %\hfill
        \includegraphics[width = 0.18 \linewidth]{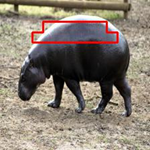}
        %\hfill
        \includegraphics[width = 0.18 \linewidth]{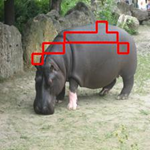}
        %\hfill
        \includegraphics[width = 0.18 \linewidth]{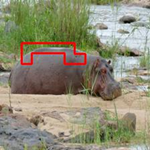}
        \caption{Hippopotamus - Torso, Back, Body top, Upper middle body, Body curve, Skin}
        \label{fig:hippopotamus_torso}
    \end{subfigure}
    \begin{subfigure}{0.5\textwidth}
        \includegraphics[width = 0.18 \linewidth]{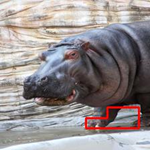}
        %\hfill
        \includegraphics[width = 0.18 \linewidth]{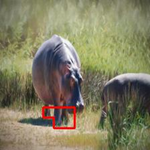}
        %\hfill
        \includegraphics[width = 0.18 \linewidth]{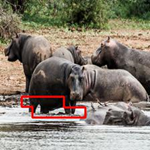}
        %\hfill
        \includegraphics[width = 0.18 \linewidth]{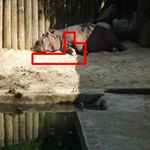}
        %\hfill
        \includegraphics[width = 0.18 \linewidth]{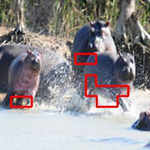}
        \caption{Hippopotamus - Legs, Limbs, Paws}
        \label{fig:hippopotamus_legs}
    \end{subfigure}
    \begin{subfigure}{0.5\textwidth}
        \includegraphics[width = 0.18 \linewidth]{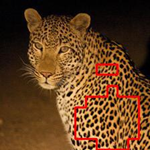}
        %\hfill
        \includegraphics[width = 0.18 \linewidth]{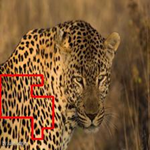}
        %\hfill
        \includegraphics[width = 0.18 \linewidth]{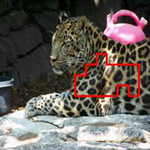}
        %\hfill
        \includegraphics[width = 0.18 \linewidth]{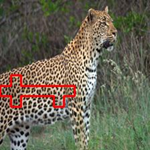}
        %\hfill
        \includegraphics[width = 0.18 \linewidth]{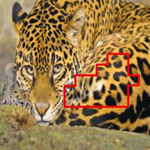}
        \caption{Leopard - Spots, Black Spots, Dots, Dot Pattern, Rosettes, Patches}
        \label{fig:leopard_spots}
    \end{subfigure}
    \begin{subfigure}{0.5\textwidth}
        \includegraphics[width = 0.18 \linewidth]{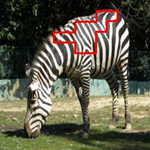}
        %\hfill
        \includegraphics[width = 0.18 \linewidth]{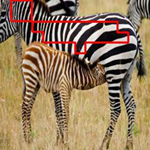}
        %\hfill
        \includegraphics[width = 0.18 \linewidth]{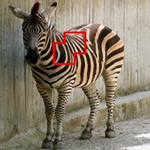}
        %\hfill
        \includegraphics[width = 0.18 \linewidth]{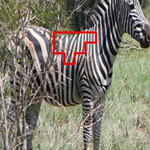}
        %\hfill
        \includegraphics[width = 0.18 \linewidth]{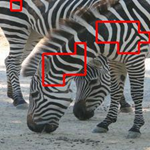}
        \caption{Zebra - Stripes}
        \label{fig:zebra_stripes}
    \end{subfigure}
    \begin{subfigure}{0.5\textwidth}
        \includegraphics[width = 0.18 \linewidth]{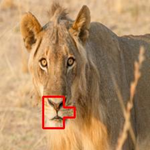}
        %\hfill
        \includegraphics[width = 0.18 \linewidth]{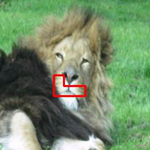}
        %\hfill
        \includegraphics[width = 0.18 \linewidth]{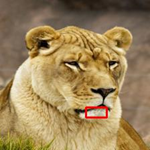}
        %\hfill
        \includegraphics[width = 0.18 \linewidth]{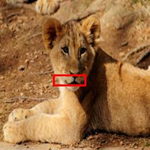}
        %\hfill
        \includegraphics[width = 0.18 \linewidth]{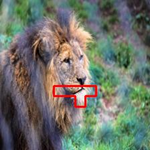}
        \caption{Lion - Mouth, Nose area}
        \label{fig:lion_mouth}
    \end{subfigure}
    \begin{subfigure}{0.5\textwidth}
        \includegraphics[width = 0.18 \linewidth]{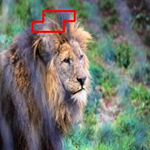}
        %\hfill
        \includegraphics[width = 0.18 \linewidth]{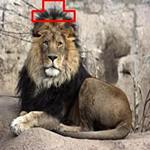}
        %\hfill
        \includegraphics[width = 0.18 \linewidth]{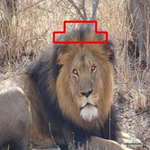}
        %\hfill
        \includegraphics[width = 0.18 \linewidth]{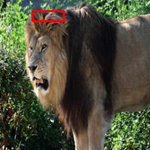}
        %\hfill
        \includegraphics[width = 0.18 \linewidth]{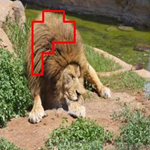}
        \caption{Lion - Mane, Top of head, Forehead hair, Upper hair on face}
        \label{fig:lion_mane}
    \end{subfigure}
    \caption{[Best viewed in color] Concept visualizations and tags given to them by the survey participants.}
    \label{fig:survey_tags}
\end{figure}

The consistency of the concepts is measured as the percentage of the participants who agreed with the presence of a common pattern across the five visualizations. The overall consistency from the experiments was observed to be 72\%. Figure ~\ref{fig:survey_bargraph} presents the class-wise consistency of the concepts. All classes except \textit{chihuahua} demonstrate high consistency. Figure ~\ref{fig:survey_tags} presents the visualizations of the concepts and the tags given by the human subjects. It can be observed that the concepts are human interpretable, as the tags are meaningful. The extracted concepts indeed are some of the features of the animals and their natural surroundings according to which humans identify these animals. 
The tags for the different concepts across the classes labeled by the participants are shown in Table ~\ref{tab:concepts_table}.
%The concept visualizations some of the for this class is shown in Figure ~\ref{fig:chihuahua_uninterpretable}. It throws light on a possible reason for the low consistency of the chihuahua class as it is a domestic animal among all other animals in our dataset. Hence the model uses the presence of household objects and dog ties, which are not found in other classes as discriminative concepts, to identify a Chihuahua. This being different from how humans perceive a chihuahua might have lead rating the concepts of the chihuahua class uninterpretable. 

% Figure ~\ref{fig:survey_tags} presents the visualizations of the concepts and the tags given by the human subjects. As can be observed, the concepts are human interpretable, as the tags are meaningful. The extracted concepts indeed are some of the features of the animals and their natural surroundings according to which humans identify these animals.
\begin{figure}
    \centering
    \begin{subfigure}{0.45\textwidth}
        \includegraphics[width = 0.18 \linewidth]{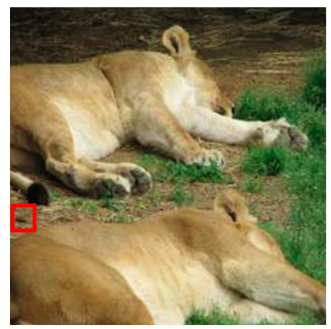}
        %\hfill
        \includegraphics[width = 0.18 \linewidth]{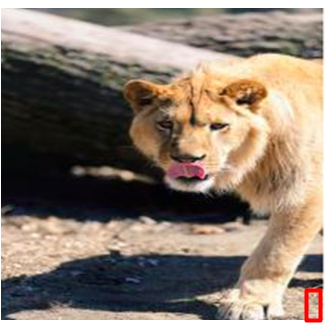}
        %\hfill
        \includegraphics[width = 0.18 \linewidth]{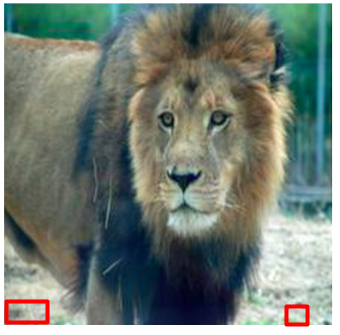}
        %\hfill
        \includegraphics[width = 0.18 \linewidth]{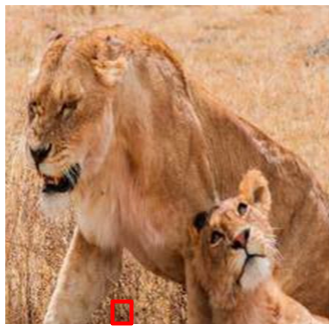}
        %\hfill
        \includegraphics[width = 0.18 \linewidth]{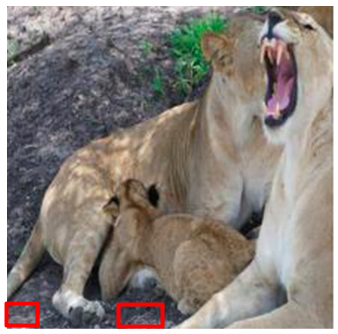}
        %\hfill
        \caption{Lion}
        \label{fig:lion_uninterpretable}
    \end{subfigure}
\begin{subfigure}{0.45\textwidth}
    \includegraphics[width = 0.18 \linewidth]{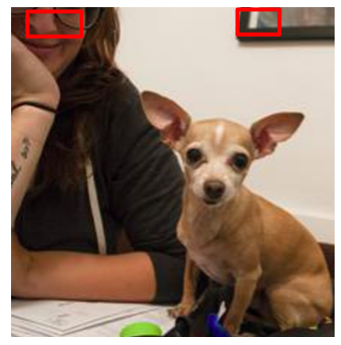}
    %\hfill
    \includegraphics[width = 0.18 \linewidth]{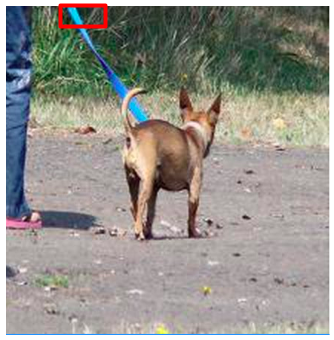}
    %\hfill
    \includegraphics[width = 0.18 \linewidth]{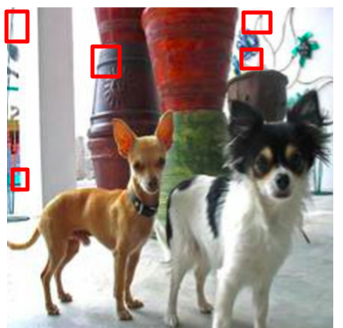}
    %\hfill
    \includegraphics[width = 0.18 \linewidth]{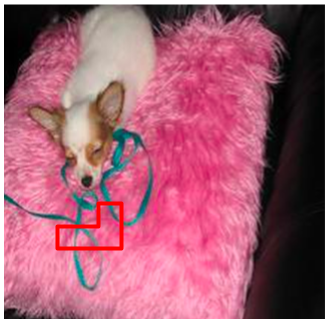}
    %\hfill
    \includegraphics[width = 0.18 \linewidth]{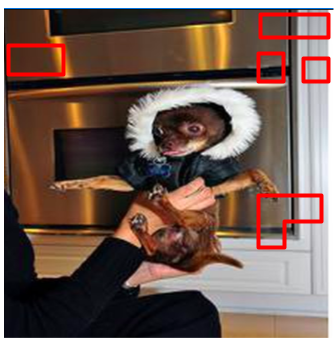}
    %\hfill
    \caption{Chihuahua}
    \label{fig:chihuahua_uninterpretable}
\end{subfigure}
\begin{subfigure}{0.45\textwidth}
    \includegraphics[width = 0.18 \linewidth]{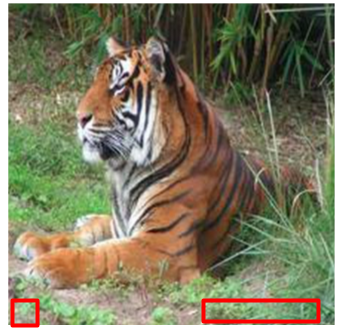}
    %\hfill
    \includegraphics[width = 0.18 \linewidth]{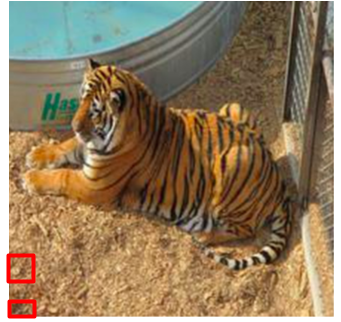}
    %\hfill
    \includegraphics[width = 0.18 \linewidth]{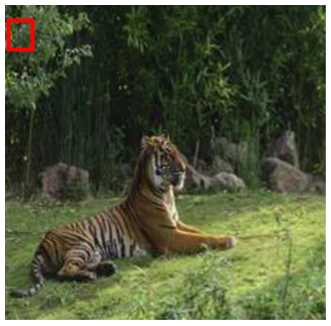}
    %\hfill
    \includegraphics[width = 0.18 \linewidth]{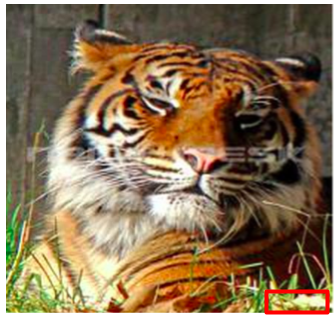}
    %\hfill
    \includegraphics[width = 0.18 \linewidth]{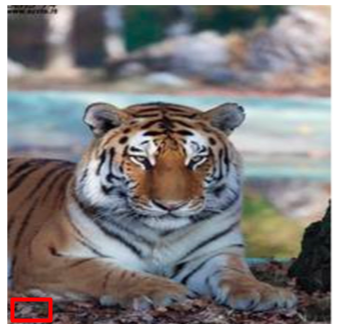}
    %\hfill
    \caption{Tiger}
    \label{fig:tiger_uninterpretable}
\end{subfigure}
    \caption{[Best viewed in color] Examples of uninterpretable concepts}
    \label{fig:uninterpretable_concepts_vis}
\end{figure}
\begin{table}
    \centering
    \begin{tabular}{|c|p{6cm}|}
        \hline
       \textbf{Class}  & \textbf{Interpretable Concept Tags} \\
       \hline
       Bobcat  & Legs, Ears, Body hair, Back, Ear hair, Grass, Ear tips, Beard\\
       \hline
       Chihuahua & Ears\\
       \hline
       Elephant & Head, Trees, Ground, Eyes, Ears, Face, Trunk, Grass, Water\\
       \hline
       Gorilla & Limbs, Forehead, Grass, Wood, Trees, Head\\
       \hline
       Hippopotamus & Legs, Feet, Water, Back, Background, Sand\\
       \hline
       Horse & Mouth, Nose, Nostrils, Mane, Ears, Grass, Hair, Neck, Back\\
       \hline
       Leopard & Mouth, Grass, Trees, Spots\\
       \hline
       Lion & Lower mane, Trees, Mouth, Back skin, Head,  Upper mane, Grass, Paws\\
       \hline
       Tiger & Paws, Ears, Legs, Background, White skin\\
       \hline
       Zebra & Stripes, Grass, Feet, Ground, Ears, Mouth\\
       \hline
    \end{tabular}
    \caption{Key concepts tagged by participants for each class}
    \label{tab:concepts_table}
\end{table}

\subsection{Qualitative Concept Analysis}
In Figure ~\ref{fig:survey_tags}, it can be seen that various concepts like ears of bobcat in Figure ~\ref{fig:bobcat_ear}, face of an elephant in Figure ~\ref{fig:elephant_face}, etc. being extracted by the PACE framework. A good visual consistency backed up by human subject votes is observed in the extracted concepts. Figure ~\ref{fig:zebra_stripes} tagged as stripes of the zebra seem to consistently highlight the stripes present in the torso region of the animal. A similar observation can be made in Figure ~\ref{fig:leopard_spots} tagged as spot patterns, the concept highlighted consistently shows the torso of the animal. This qualitatively shows that consistent concept embeddings have been learned as expected.

%We also see a few concepts that were marked uninterpretable by the human subjects presented in Figure ~\ref{fig:uninterpretable_concepts_vis}. Figure ~\ref{fig:lion_uninterpretable} seems to highlight sand dirt around the legs of the lion and Figure ~\ref{fig:tiger_uninterpretable} highlights grass around the tiger. As it can be seen that the area highlighted to depict the concept itself is very small, only participants with greater attention to details were able to tag such concepts. The majority of the participants deemed it to be uninterpretable. We can associate the detection of uninterpretable concepts from the feature maps to the residuals extracted from the feature map during matrix factorization based explanation techniques \cite{ice}. This also proves the effectiveness of PACE that a good approximation of the internals in the feature map has been extracted through interpretable (conceptually analogous to factors in matrix factorization \cite{ice}) and uninterpretable concepts (conceptually analogous to residuals in matrix factorization \cite{ice}).
A few concepts that were marked uninterpretable by the human subjects is presented in Figure ~\ref{fig:uninterpretable_concepts_vis}. Figure ~\ref{fig:lion_uninterpretable} seems to highlight sand dirt around the legs of the lion and Figure ~\ref{fig:tiger_uninterpretable} highlights grass around the tiger. As it can be seen that the area highlighted to depict the concept itself is very small, only participants with greater attention to details were able to tag such concepts. The majority of the participants deemed it to be uninterpretable. The detection of uninterpretable concepts from the feature maps can be associated to the residuals extracted from the feature map during matrix factorization based explanation techniques \cite{ice}. This also proves the effectiveness of PACE that a good approximation of the internals in the feature map has been extracted through interpretable (conceptually analogous to factors in matrix factorization \cite{ice}) and uninterpretable concepts (conceptually analogous to residuals in matrix factorization \cite{ice}).

\begin{figure}
    \centering
    \begin{subfigure}{0.5 \textwidth}
        \includegraphics[width = 0.18 \linewidth]{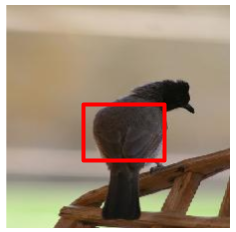}
        %\hfill
        \includegraphics[width = 0.18 \linewidth]{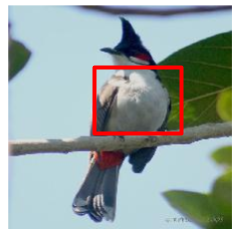}
        %\hfill
        \includegraphics[width = 0.18 \linewidth]{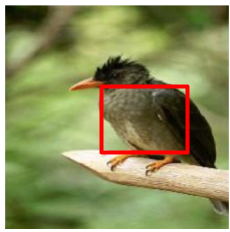}
        %\hfill
        \includegraphics[width = 0.18 \linewidth]{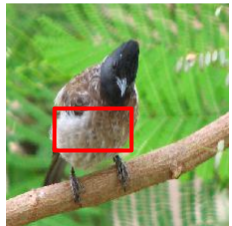}
        %\hfill
        \includegraphics[width = 0.18 \linewidth]{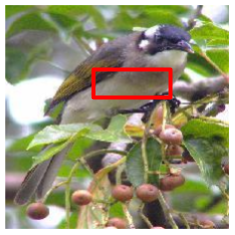}
        %\hfill
        \caption{Bulbul - Body}
        \label{fig:bulbul_body}
    \end{subfigure}
    \begin{subfigure}{0.5 \textwidth}
        \includegraphics[width = 0.18 \linewidth]{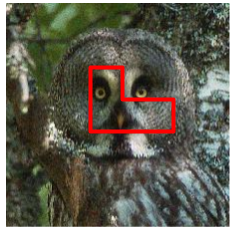}
        %\hfill
        \includegraphics[width = 0.18 \linewidth]{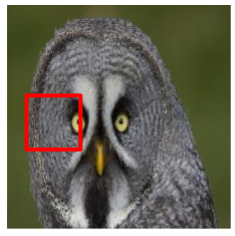}
        %\hfill
        \includegraphics[width = 0.18 \linewidth]{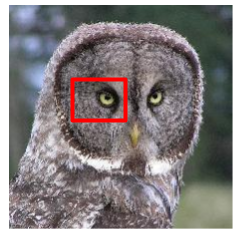}
        %\hfill
        \includegraphics[width = 0.18 \linewidth]{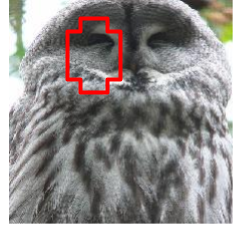}
        %\hfill
        \includegraphics[width = 0.18 \linewidth]{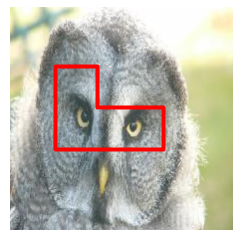}
        %\hfill
        \caption{Great Grey Owl - Eyes}
        \label{fig:owl_eyes}
    \end{subfigure}
    \begin{subfigure}{0.5 \textwidth}
        \includegraphics[width = 0.18 \linewidth]{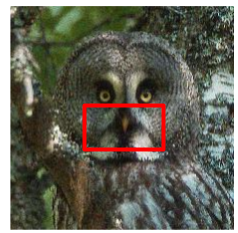}
        %\hfill
        \includegraphics[width = 0.18 \linewidth]{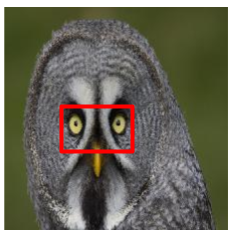}
        %\hfill
        \includegraphics[width = 0.18 \linewidth]{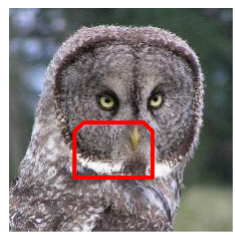}
        %\hfill
        \includegraphics[width = 0.18 \linewidth]{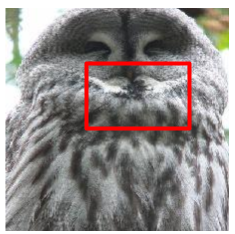}
        %\hfill
        \includegraphics[width = 0.18 \linewidth]{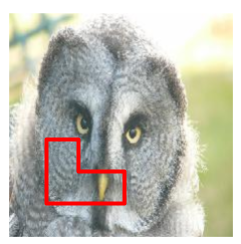}
        %\hfill
        \caption{Great Grey Owl - Beak}
        \label{fig:owl_beak}
    \end{subfigure}
    \begin{subfigure}{0.5\textwidth}
        \includegraphics[width = 0.18 \linewidth]{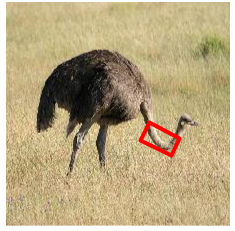}
        %\hfill
        \includegraphics[width = 0.18 \linewidth]{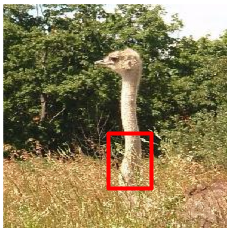}
        %\hfill
        \includegraphics[width = 0.18 \linewidth]{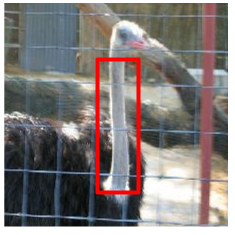}
        %\hfill
        \includegraphics[width = 0.18 \linewidth]{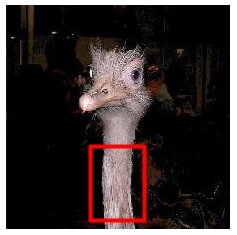}
        %\hfill
        \includegraphics[width = 0.18 \linewidth]{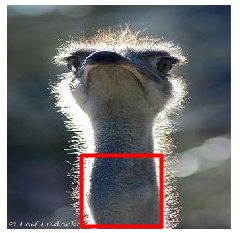}
        %\hfill
        \caption{Ostrich - Neck}
        \label{fig:ostrich_neck}
    \end{subfigure}
    \begin{subfigure}{0.5\textwidth}
        \includegraphics[width = 0.18 \linewidth]{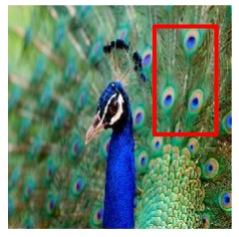}
        %\hfill
        \includegraphics[width = 0.18 \linewidth]{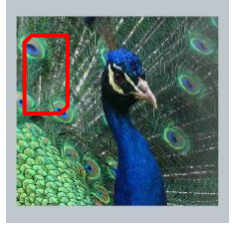}
        %\hfill
        \includegraphics[width = 0.18 \linewidth]{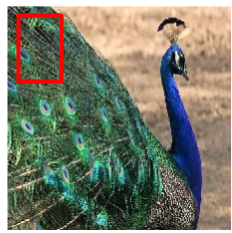}
        %\hfill
        \includegraphics[width = 0.18 \linewidth]{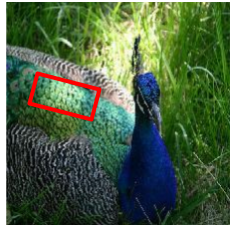}
        %\hfill
        \includegraphics[width = 0.18 \linewidth]{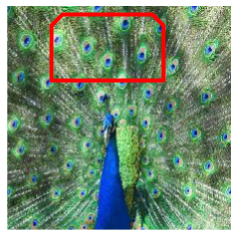}
        %\hfill
        \caption{Peacock - Feathers}
        \label{fig:peacock_feather}
    \end{subfigure}
    \begin{subfigure}{0.5\textwidth}
        \includegraphics[width = 0.18 \linewidth]{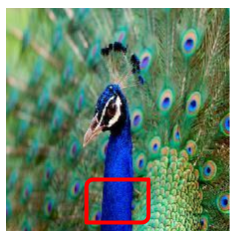}
        %\hfill
        \includegraphics[width = 0.18 \linewidth]{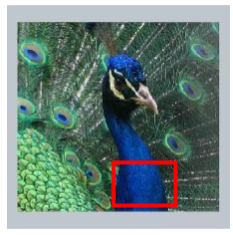}
        %\hfill
        \includegraphics[width = 0.18 \linewidth]{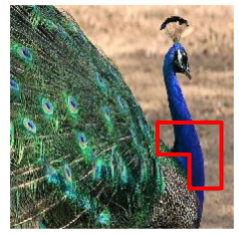}
        %\hfill
        \includegraphics[width = 0.18 \linewidth]{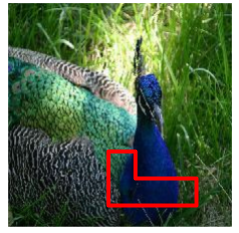}
        %\hfill
        \includegraphics[width = 0.18 \linewidth]{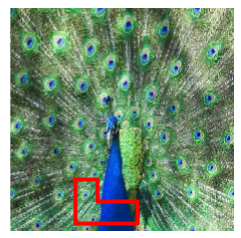}
        %\hfill
        \caption{Peacock - Neck}
        \label{fig:peacock_neck}
    \end{subfigure}
    \begin{subfigure}{0.5\textwidth}
        \includegraphics[width = 0.18 \linewidth]{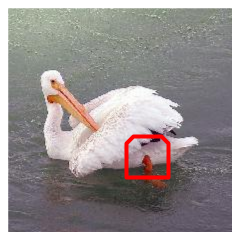}
        %\hfill
        \includegraphics[width = 0.18 \linewidth]{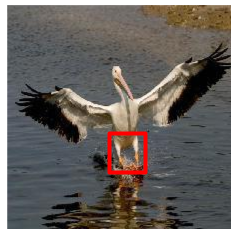}
        %\hfill
        \includegraphics[width = 0.18 \linewidth]{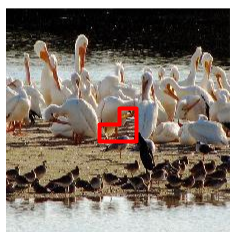}
        %\hfill
        \includegraphics[width = 0.18 \linewidth]{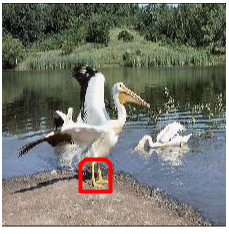}
        %\hfill
        \includegraphics[width = 0.18 \linewidth]{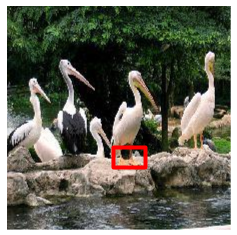}
        %\hfill
        \caption{Pelican - Legs}
        \label{fig:pelican_legs}
    \end{subfigure}
    \begin{subfigure}{0.5 \textwidth}
        \includegraphics[width = 0.18 \linewidth]{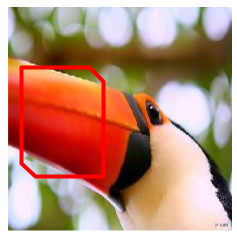}
        %\hfill
        \includegraphics[width = 0.18 \linewidth]{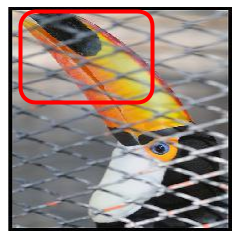}
        %\hfill
        \includegraphics[width = 0.18 \linewidth]{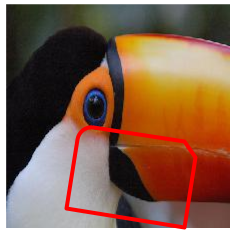}
        %\hfill
        \includegraphics[width = 0.18 \linewidth]{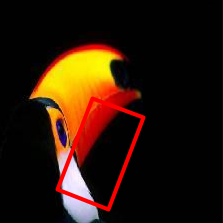}
        %\hfill
        \includegraphics[width = 0.18 \linewidth]{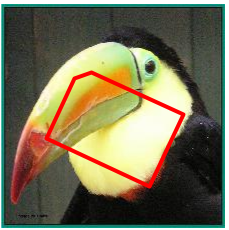}
        %\hfill
        \caption{Toucan - Bill}
        \label{fig:toucan_bill}
    \end{subfigure}
    \begin{subfigure}{0.5 \textwidth}
        \includegraphics[width = 0.18 \linewidth]{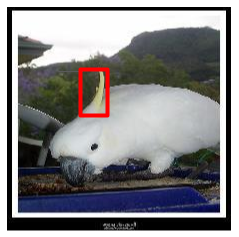}
        %\hfill
        \includegraphics[width = 0.18 \linewidth]{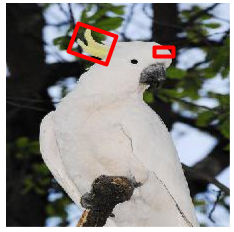}
        %\hfill
        \includegraphics[width = 0.18 \linewidth]{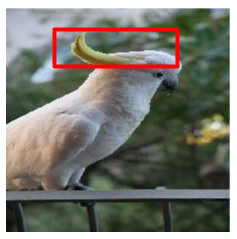}
        %\hfill
        \includegraphics[width = 0.18 \linewidth]{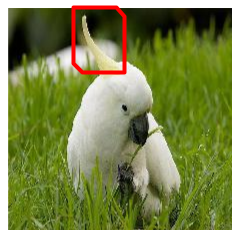}
        %\hfill
        \includegraphics[width = 0.18 \linewidth]{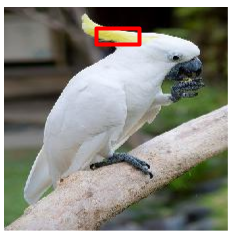}
        %\hfill
        \caption{Sulphur Crested Cockatoo - Crest}
        \label{fig:cockatoo_crest}
    \end{subfigure}
    \begin{subfigure}{0.5 \textwidth}
        \includegraphics[width = 0.18 \linewidth]{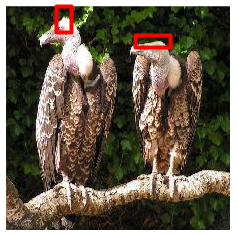}
        %\hfill
        \includegraphics[width = 0.18 \linewidth]{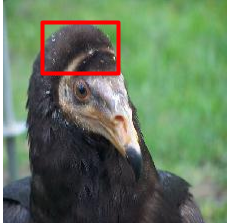}
        %\hfill
        \includegraphics[width = 0.18 \linewidth]{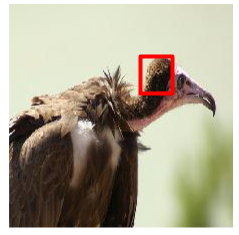}
        %\hfill
        \includegraphics[width = 0.18 \linewidth]{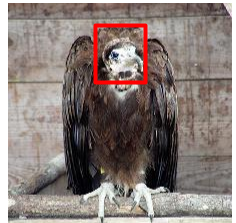}
        %\hfill
        \includegraphics[width = 0.18 \linewidth]{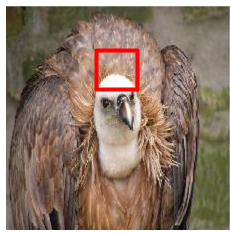}
        %\hfill
        \caption{Vulture - Head}
        \label{fig:vulture_head}
    \end{subfigure}
    \caption{[Best viewed in color] Visualization of the concepts extracted from VGG19 model trained on Imagenet-Birds dataset}
    \label{fig:vgg19_birds_vis}
\end{figure}

Figure ~\ref{fig:vgg19_birds_vis} shows the concepts extracted by PACE for the VGG19 black-box model trained on the Imagenet-Birds dataset. Salient parts of the birds like feathers of a peacock in Figure ~\ref{fig:peacock_feather}, blue neck in Figure ~\ref{fig:peacock_neck}, eyes of Great Grey Owl in Figure ~\ref{fig:owl_eyes}, its beak in Figure ~\ref{fig:owl_beak}, toucan's characteristic colorful bill in Figure ~\ref{fig:toucan_bill}, crest of Sulphur Crested Cockatoo in Figure ~\ref{fig:cockatoo_crest}, etc. seems to be detected by PACE. These parts are indeed discriminatory features that help distinguish the particular bird species from other bird species. A good visual consistency can also be found in the concepts visualized across different images.
%We can see salient parts of the birds like feathers of a peacock in Figure ~\ref{fig:peacock_feather}, blue neck in Figure ~\ref{fig:peacock_neck}, eyes of Great Grey Owl in Figure ~\ref{fig:owl_eyes}, its beak in Figure ~\ref{fig:owl_beak}, toucan's characteristic colorful bill in Figure ~\ref{fig:toucan_bill}, crest of Sulphur Crested Cockatoo in Figure ~\ref{fig:cockatoo_crest}, etc. being detected by PACE. These parts are indeed discriminatory features that help distinguish the particular bird species from other bird species. We can also find good visual consistency in the concepts visualized across different images.

\subsection{Explaining Misclassifications}
\begin{figure}
    \centering
    \begin{subfigure}[b]{0.15\textwidth}
            \centering
             \includegraphics[width= 0.7 \linewidth]{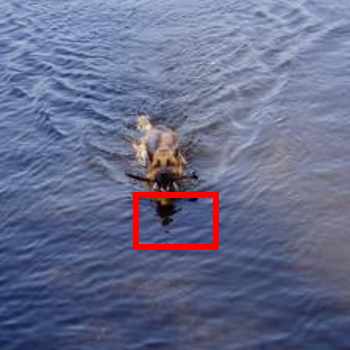}
               \caption{}
              \label{fig:germanshepherd_as_hippo}
        \end{subfigure}
        %\hfill
        \begin{subfigure}[b]{0.15\textwidth}
            \centering
             \includegraphics[width= 0.7   \linewidth]{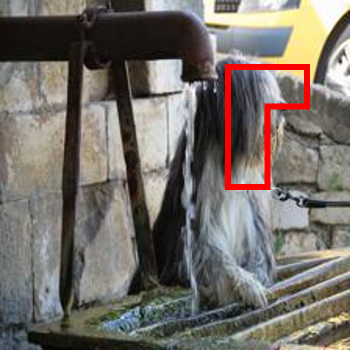}
               \caption{}
              \label{fig:collie_as_horse}
        \end{subfigure}
        %\hfill
        \begin{subfigure}[b]{0.15\textwidth}
            \centering
             \includegraphics[width= 0.7   \linewidth]{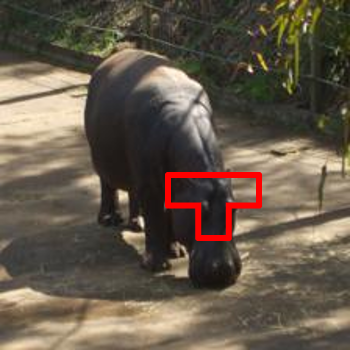}
               \caption{}
              \label{fig:hippo_as_elephant}
        \end{subfigure}
    \caption{[Best viewed in color] Misclassified Images - (a) German Shepherd misclassified as Hippopotamus, (b) Collie misclassified as Horse, (c) Hippopotamus misclassified as Elephant}
    \label{fig:misclassified_images}
\end{figure}

The class-discriminative concepts learned by PACE can be used to explain black-box model misclassifications. Figure ~\ref{fig:misclassified_images} presents a few examples of misclassified images and their salient concepts extracted by PACE. Figure ~\ref{fig:germanshepherd_as_hippo} shows an image of a \textit{german shepherd} misclassified as a \textit{hippopotamus}. Understandably, the model uses the concept of water specific to the \textit{hippopotamus} class for this prediction. Similarly, Figure ~\ref{fig:collie_as_horse} shows a \textit{collie} being misclassified as a \textit{horse} due to high support from the concept corresponding to the mane of the horse and Figure ~\ref{fig:hippo_as_elephant} shows a \textit{hippopotamus} misclassified as an \textit{elephant} due to high support from the concept corresponding to the head of the elephant. The explanations show that the model is wrong for the right reasons.

%\section{Summary}
\section{Conclusion} % as suggested by Reviewer 1
%We propose the PACE framework that learns to extract class-specific concepts relevant to the black-box prediction. The relevance is formulated such that the explanations are faithful to the black-box prediction by design. We experiment with our explainer's applicability on datasets like AWA2 and Imagenet-Birds as well on black-box architectures like VGG16 and VGG19. Qualitative and quantitative analyses show that PACE extracts concepts that are consistent and relevant. Extensive human subject experiments show that our framework provides interpretable concepts.
The PACE framework that learns to extract class-specific concepts relevant to the black-box prediction is proposed.The relevance is formulated such that the explanations are faithful to the black-box prediction by design. The explainer's applicability on datasets like AWA2 and Imagenet-Birds as well on black-box architectures like VGG16 and VGG19 is experimented. Qualitative and quantitative analyses show that PACE extracts concepts that are consistent and relevant. Extensive human subject experiments show that the proposed framework provides interpretable concepts.

\section*{Acknowledgment}
The resources provided by ‘PARAM Shivay Facility’ under the National Supercomputing Mission, Government of India at the Indian Institute of Technology, Varanasi, and under Google Tensorflow Research award are gratefully acknowledged.

\bibliographystyle{IEEEtran}
\bibliography{IEEEabrv,ref}

\end{document}